\documentclass[twocolumn,letterpaper,10pt]{article}

\usepackage{times}
\usepackage{epsfig}
\usepackage{epstopdf}
\usepackage{graphicx}
\usepackage{amsmath}
\usepackage{amssymb}
\usepackage{amsfonts,mathrsfs,amscd,amsthm,xcolor}
\usepackage{enumitem}
\usepackage{hyperref}

\title{Manifold Hypothesis in Data Analysis: Double Geometrically-Probabilistic Approach to Manifold Dimension Estimation}

\author{Alexander Ivanov, Gleb Nosovskiy, Alexey Chekunov, Denis Fedoseev, \\ Vladislav Kibkalo, Mikhail Nikulin, Fedor Popelenskiy, Stepan Komkov, \\ Ivan Mazurenko, Aleksandr Petiushko}

\theoremstyle{definition}

\def\eg{\emph{e.g}.\,}

\def\etal{\emph{et al}.\,}
\def\ie{\emph{i.e}.\,}

\date{}
\begin{document}

\maketitle

\begin{abstract}
Manifold hypothesis states that data points in high-dimensional space actually lie in close vicinity of a manifold of much lower dimension. In many cases this hypothesis was empirically verified  and used to enhance unsupervised and semi-supervised learning. Here we present new approach to manifold hypothesis checking and underlying manifold dimension estimation. In order to do it we use two very different methods simultaneously --- one geometric, another probabilistic --- and check whether they give the same result. Our geometrical method is a modification for sparse data of a well-known box-counting algorithm for Minkowski dimension calculation. The probabilistic method  is new. Although it exploits standard nearest neighborhood distance, it is  different from methods which were previously used in such situations. This method is robust, fast and includes special preliminary data transformation. Experiments on real datasets show that the suggested approach based on two methods combination is powerful and effective.
\end{abstract}

\section{Introduction}

In recent years a lot of results were obtained in the intersection of three branches of mathematics: Neural Network (NN) theory, statistics and geometry \cite{Nonlin,LocPCA,dimRed,Anomaly}. It turns out that geometrical and topological approaches can be used for more precise and effective description of datasets (\eg output of some NN) \cite{Burn,Grassm1,Riem2} and for NN training \cite{Grassm3}. 
These approaches exploit the following {\it  Manifold Hypothesis:} non-artificial datasets in high-dimensional space often lie in a neighborhood of some manifold (surface) of much smaller dimension \cite{Fefferman}. 

The paper is devoted to the problem of estimating the dimension of this manifold. This problem is important because several effective geometrical approaches and algorithms of datasets processing require preliminary knowledge of manifold dimension \cite{Fefferman,Burn}. Our approach is to use two very different methods simultaneously --- one geometric, another probabilistic. Finally we check whether both give the same result. The first method is a modification for sparse data of a well-known box-counting algorithm for Minkowski dimension calculation. The second one is new. Although it exploits standard nearest neighborhood distance, it is  different from methods which were previously used in such situations.

Let start with a brief survey of existing approaches to manifold problem in data analysis.

One of the first results in dataset dimension reduction was obtained by Karl Pearson basing on Principal Components Analysis (PCA) \cite{Pearson01}. But PCA requires that dataset is distributed near some linear manifold, otherwise this approach is not effective.

Fefferman \etal in \cite{Fefferman} suggest an algorithm which for a given dataset of an unknown nature chooses between two possibilities: (a) there exist a ``bad'' approximating manifold, and (b) there is no ``good'' approximating manifold. By ``good'' and ``bad'' we mean the following: the better is the manifold, the less is its volume and the greater is its neighbourhood  where the projection on it is unique (manifold {\it reach}). This algorithm has exponentially growing complexity and is not applicable for huge datasets with several million samples. See also \cite{EstReach}. This method was  further developed in  \cite{Feffer3}, \cite{Feffer2}. In \cite{Feffer2} reconstruction problem for a manifold with problem of manifold embedding to linear space of smaller dimension. Problem of good embedding construction was discussed in \cite{Burn}. Grassmann-Stiefel Eigenmaps approach is compared there with several other methods (using Swissroll dataset as example) and is shown to be rather effective. Grassmannian space approach was also used in \cite{Grassm1}.
This approach was further developed and used for convolutional NN-training and explicit reconstruction of a 2-manifold in 3-space \cite{Grassm3}. 

Problem of manifold intrinsic dimension estimating arose, for example, in the context of neuro-biological studies, see \cite{AtSoPe}; that paper also contains a short survey of popular methods of the dimension estimation. In \cite{MorMed} an attempt was made to work in the ambient space directly without preliminary dimension reduction. There are two most common approaches to dimension estimation: {\em projective} and {\em geometric}. 

The idea of projective approach is to look at ``jumps'' in the magnitude of  sorted eigenvalues of the covariance matrix of the set of points. Idea is that abrupt change of eigenvalues should occur approximately between $d$-th and $(d+1)$-th sorted eigenvalues of $D\times D$ covariance matrix. But  threshold for significient ``jump'' depends on the data and cannot be reliably chosen without some knowledge about the  manifold.

This method has two main disadvantages. First, while being pretty effective in linear case (\ie when  manifold $M^d$ is  a linear space $\mathbb{R}^d$ somehow embedded into the space $\mathbb{R}^D$), it often fails on manifolds with high intrinsic curvature. To some extent this problem  can be solved by multiscale analysis. The second disadvantage is that this method requires at least $N\sim d\log d$ data points. For smaller  datasets it does not give any definite answer. It makes it pretty ineffective for undersampled data on even a moderately high-dimensional manifold. 

The geometric approach originally aroused in fractal dimension study of strange attractors in dynamical systems \cite{Grassberger}. The idea is to look at the so-called {\em correlation integral}
$\rho_X(r)=\frac{2}{N(N-1)}\sum_{1\le n<m\le N} \theta(r-||X_n-X_m||)$
for a given cutoff radius $r$. It measures the density of neighbours of a point within the $r$-ball around it. It was observed that with $r\to 0$  correlation integral approaches $r^d$, which allows to estimate $d$ looking at the slope of the linear part of $\rho_X(r)$ function in the appropriate logarithmic scale. This approach is very effective when dimension $d$ is small (up to 10 in general). For higher dimensions the method fails, underestimating the dimension dramatically, see \cite{ER}. 

Both approaches lay upon the assumption that locally data points are distributed uniformly in a $d$-dimensional ball linearly embedded in $\mathbb{R}^D$. It was addressed in \cite{Erba}, where V.~Erba \etal \cite{Erba} propose a  novel approach which suggests to look at the {\em average} correlation integral for the boundary of  $d$-dimensional disc of the radius $r_s$ one the manifold. It is defined by the formula
${\bar \rho_S({\bar r})}=\frac{1}{2}+\frac{\Omega_{d-1}}{\Omega_d}({\bar r}-2)F_{2,1}(\{\frac{1}{2}, 1-\frac{d}{2}\},\{\frac{3}{2}\} ||({\bar r}-2)^2),$
where ${\bar r}=\frac{r}{r_s}$, $\Omega_d$ is the $d$-dimensional solid angle, and $F_{2,1}$ is the $(2,1)$-hypergeometric function. This formula was taken as the definition of the {\em full correlation integral} (FCI). For highly curved manifolds, a multiscale modification was developed which takes in account  differences in local geometry of the manifold. To our knowledge this method was not applied to real (non-artificial) dataset and apart from the case of two linearly embedded linear spaces, it was not tested on manifolds of high dimension $d$. For dimension $d=15$ it gives  broad  estimation interval with only its left edge corresponding to the actual dimension value. 

Other methods of dimension estimation include the use of approximating geodesics with minimum distance paths on a graph (see \cite{GraCar}), by taking the number of samples that fall into a ball radius of which is calculated by graph distance (which is approximation of geodesic distance on manifold), and estimating the dimension on each sample (see \cite{HeJi}). An interesting paper \cite{MaxLike} contains a method of the estimation of intrinsic dimension of a manifold approximating a data cloud, by looking at the distances between close neighbours and using the properties of Poisson processes. This statistics-based approach resembles the dimension estimation method we propose in the present paper.

Let us conclude this short overview of different known methods by emphasising that the methods we describe and propose in the present paper are suggested for a particular (and very complex) type of datasets, where the samples are distributed uniformly in all the ambient space (and along the manifold as well), and the manifold consists of closely situated leaves, each of them spreading along most of the ambient space. The methods we are aware of do not give good estimates of the intrinsic dimension for that intriguing case of datasets.

\section{New Synthetic Approach to Dimension Estimation}

In present section we suggest  new dimension estimation approach  based on combination of two quite different methods: {\em Minkowski dimension} calculation and {\em geometrically-probabilistic method} based on the nearest neighbour distance distribution after a special data transformation which we call "flattening".

\subsection{Minkowski dimension}

The first method is purely geometrical. The idea is that tiling Euclidean space with small cubes and counting the number of cubes required to cover a given subset gives us an asymptomatic description of  fractal dimension of this set --  {\em Minkowski dimension} \cite{Falconer}. Minkowski dimension is equal to manifold dimension if our set is a manifold.  Well-known {\em box-counting algorithm} can be used for its calculation. 

Minkowski dimension of a subset $M \subset \mathbb{R}^m$ is defined as follows. Let us split $\mathbb{R}^m$ into the union of distinct $m$-dimensional cubes with side $r$, and denote the set of those cubes by $B_r$. Consider the number $N(r)$ of cubes which intersect the set $M$:
$	N(r) = |\{B\in B_r \,|\, B\cap M\neq \emptyset\}|.
$
{\em Minkowski dimension} of the set $M$ is the limit
$
		\dim_{mink}(M)=\lim\limits_{r\to 0}\frac{\log N(r)}{\log(1/r)}.
$

\begin{figure}[!t]
\centering\includegraphics[width=1\linewidth]{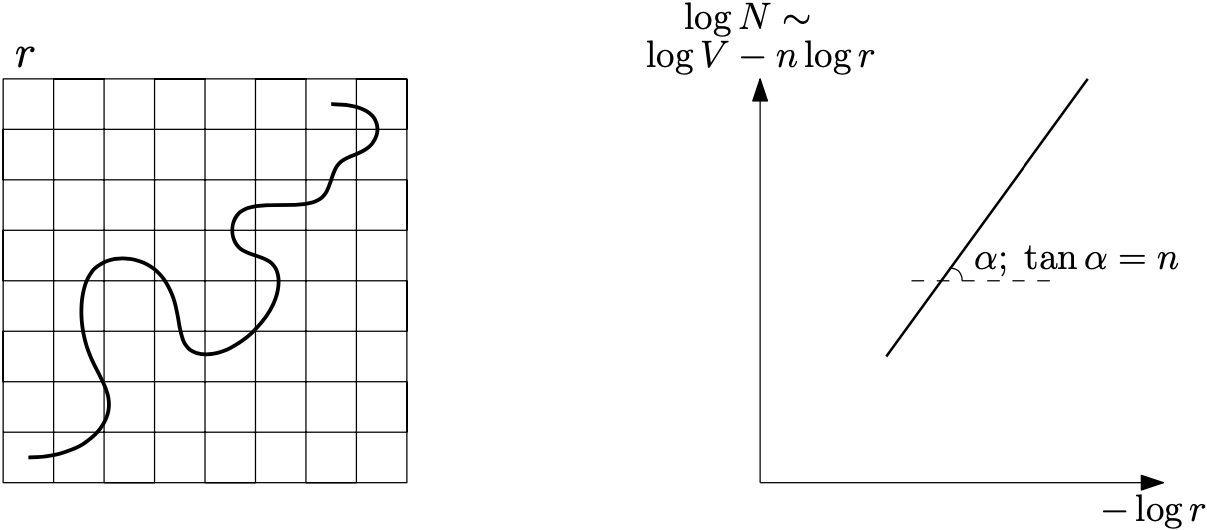}
\caption{Minkowski dimension for a 1-dimensional curve in a 2-dimensional plane. Angular coefficient: $\tan \alpha = 1$}
\label{fig:minkowski_for_manifold}
\end{figure}

 In the case when  $M$ is an $n$-dimensional manifold, as $r\to 0$ we have  $N \sim \frac{V}{r^n}$, where $V$ is the $n$-dimensional volume of $M$. Therefore, as $r\to 0$ the asymptotic holds:
\begin{equation}
	\log{N} \sim \log{V} - n \log{r}.
\label{eq:minkowski_asymptotic}
\end{equation}
It means that $\log{N}$ is a linear function from $\log{r}$. Therefore, dimension $n$ of the manifold $M$ can be found as the angular coefficient of this linear function, see Fig.~\ref{fig:minkowski_for_manifold}.

\begin{figure}[t]
\centering\includegraphics[width=1\linewidth]{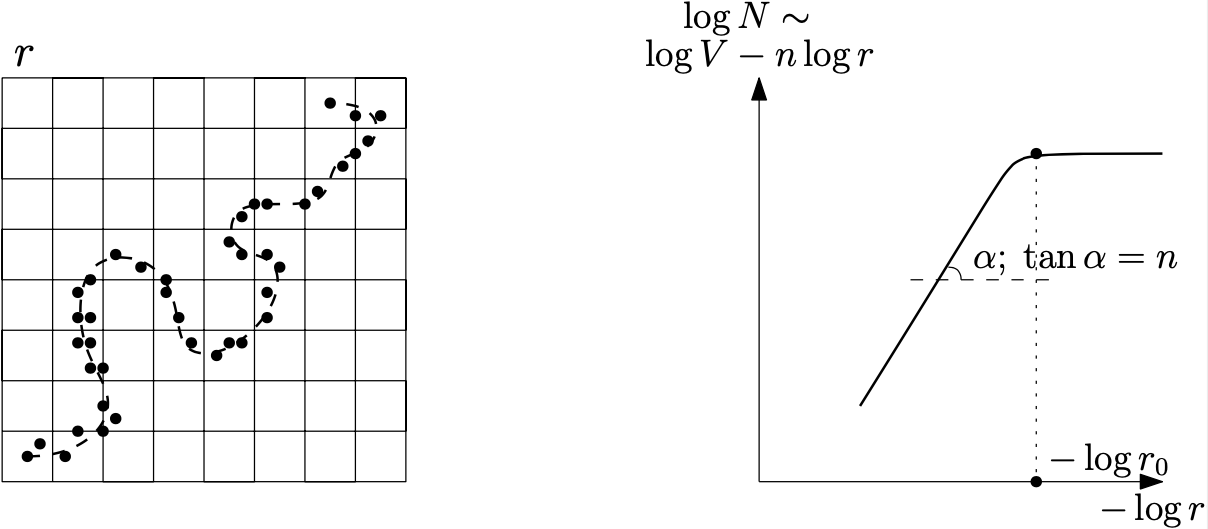}
\caption{Minkowski dimension for a cloud of points in a neighbourhood of a 1-dimensional curve in a 2-dimensional plane. For $-\log r \ge -\log r_0$ the curve becomes horizontal}
\label{fig:minkowski_for_cloud}
\end{figure}

Now let us consider a finite  cloud of data points $\mathcal{C}=\{P_i\}$ located sufficiently close to some $n$-dimensional manifold $M$. We want to find dimension $n$ of $M$ applying the box counting algorytm to the cloud $\mathcal{C}$. Now let us define by $N(r)$
 number of cubes which contain points from $\mathcal{C}$.
Assuming $C$ contains sufficiently large number of points (for certain $m$ and $n$), we could estimate $n$ by the asymptotic formula (\ref{eq:minkowski_asymptotic}). 
Note that formula (\ref{eq:minkowski_asymptotic}) for $\log{N}$ holds only until $r > r_0 > \frac{d}{\sqrt{m}}$ for some $r_0$, where $d$ is the minimal distance between points in our cloud: $d=\min_{i\neq j} \{|P_i - P_j|\}$. Indeed, if $r < \frac{d}{\sqrt{m}}$ then each cube contains not more then one data point. Hence for $r < \frac{d}{\sqrt{m}}$ we have: $\log{N}(r) =\log |\mathcal{C}|= \text{const}$,
see Fig.~\ref{fig:minkowski_for_cloud}. It means that in this case dimension should be estimated using values $r > r_0$ only.

We are mostly interested in the case when $m$ is very  large, for example $m=512$. In this case, the threshold $r_0$ will be far from $0$, and it cannot be changed by increasing the size of  $\mathcal{C}$. Indeed, let us consider a cloud  $\mathcal{C}$ located in the cube with edge $2a$. If we divide this cube into half-size cubes with edge $a$, then the number of these $a$-cubes will be $2^{512}$, which exceeds the number of atoms in the observable universe. It means, that for any real $\mathcal{C}$ with more or less uniform distribution,  almost all of these $a$-cubes will  contain not more than one point from $\mathcal{C}$. Hence, $r_0>a$. 

However, as we observed in experiments, in many cases of practical interest  in the range $r >r_0$ there still can be observed an interval of linear behaviour of the function $\log{N} =F(-\log{r}),$ large enough to determine dimension $n$, see Fig.~\ref{fig:minkowski_for_cloud}.
There are some nontrivial theoretical arguments supporting this observation, but discussion of therm is out of scope of this work.

Note, that
	if the dimension of the ambient space is large, then in order to count occupied cubes one should step over points of the dataset and look at cubes occupied by them, rather then considering all cubes, which is practically impossible.

\subsection{Geometrically-probabilistic Method}

The second method combines geometrical and probabilistic ideas to estimate the  dimension.

Consider a finite set $P \subset \mathbb{R}^m$ of independent uniformly distributed random points in $\mathbb{R}^m$. Let us suppose that  $P$ belongs to some $n$-dimensional manifold $M^n\subset \mathbb{R}^m$. For each $x \in P$ we can define its closest neighbor distance as $d_{min}(x) = \min \{|x-y|\colon y \neq x, \, y$

It is well known that under our assumptions the distribution of $d_{min}$ is such that $V^n(d_{min})$ has exponential distribution with some intensity parameter $\lambda$: $f(t;n) dt= \lambda e^{- \lambda V^n(t)} dV^n(t),$ where $V^n(t)$ stands for the $n$-dimensional volume or the ball of radius $t$ \cite{Kendall}.  The value of $V^n(t)$ is a well-known function of $n$ and $t$.

Our algorithm of dimension estimation for the manifold $M$ is as follows.

\begin{enumerate}[noitemsep,topsep=0pt]
    \item Check independence of points in the dataset. If we allow the dataset to have dependent points (which are closer to each other than  independent ones), then real dimension value will be underestimated by this method.

    \item  Uniformity of the distribution. The dataset should be uniformly distributed over the underlying manifold, and this manifold should be locally more or less close to Euclidean space (otherwise above mentioned probability features can be invalid). But we can not check it directly because  manifold is unknown to us. However, for certain face  recognition Neural Networks datasets distribution of data points (``embeddings'') in ambient space appear to be almost uniform, and therefore it is reasonable to assume that they are also uniformly distributed  over the underlying manifold (otherwise this manifold should be very specific). For such datasets ambient space is a ball of very high dimension (512 in our case), so almost all uniformly distributed data points must lay in close vicinity of the  boundary sphere (511 dimensional sphere in our case) and distribution of distances between different points must have certain form. All these features can be checked. Next we check that data points distribution in ambient space is rotational symmetric -- which implies that this distribution is uniform on the boundary sphere. In practice, to check the symmetry we choose randomly several hundred thousands of  directions from the ball center and check if all projections of our dataset to these directions have approximately the same distribution $F(x).$
    \item Flattening procedure. When we calculate  distances between data points and their nearest neighbors, we should do it along the underlying manifold. But under the necessity we  use Euclidean distance in ambient space instead. To guarantee that these two distances are close to each other, we need our manifold to be locally  flat (more or less). In order to achieve it, we use special {\it flattening} procedure. Denote by $F(x)$ common distribution of our dataset points along any coordinate axis (see previous step). We perform the following {\it flattening} coordinate transformation $x \rightarrow x': x^{i'} = F(x^i), \quad 1 \leq i,i' \leq D.$ After such transformation distribution along all new coordinates will be uniform on $[0,1]$. Boundary $(D-1)$ dimensional sphere with our dataset in its vicinity will be part-wise flattened, see Fig.~\ref{fig:flattening}.

    
	\item Fix the value of $n$ (the dimension candidate).

    \item For the chosen $n$ calculate experimental distribution of $V^n(d_{min})$.
    	
	\item Check if the obtained experimental distribution is exponential. If not, then increase n and repeat steps 3--5.
\end{enumerate}

To perform the last step we calculate the following values:
\begin{itemize}[noitemsep,topsep=0pt]
    \item {\bf Mean}: $A_1(n) = \mathbb{E}(V^n(d_{\min}))$,

    \item {\bf Variance}: $A_2(n)=Var(V^n(d_{\min}))$,

    \item {\bf Kolmogorov-Smirnov (K-S) statistic}: $D_{n} = \sup{|F_{n}(V^n(d_{min})) - F(V^n(d_{min}))|}$,  where $F_{n}(x)$ is empirical cumulative distribution function, and $F(x)$ is the theoretical one.
\end{itemize}

Let us recall that for exponential distribution $A_1^2= A_2$. We perform calculations for each $n$ and look for the situation when simultaneously $ D_{n} \rightarrow \min$ and $A_1^2 = A_2$. If such $n$ exists, we take it as the dimension estimation.

Note that the paper \cite{BeGoRaSha} states that in many cases in high dimension the distance to the nearest neighbour of a point is almost exactly the same as the distance to the farthest neighbour. This observation does not render the method proposed above incorrect, and we do not study whether the case we are dealing with satisfies the conditions of the main theorem of \cite{BeGoRaSha}.

\section{Experiments on classical setups} 

\subsection{Experiment setup}
Both proposed methods are applied to the following artificial test datasets: 
    	\begin{enumerate}
    		\item The Swiss Roll dataset $\mathcal{SR}_{2,3}$; a uniform sampling from $[0,1]^2$, embedded into $\mathbb{R}^3$ with the map 
    	$\varphi(x,y)=(x\cos{2\pi y}, y, x\sin{2\pi y}).$
			\item Linear datasets $\mathcal{H}_{3,30}$ and $\mathcal{H}_{4,30}$: uniform samplings from $[0,1]^3$ and $[0,1]^4$, linearly embedded into $\mathbb{R}^{30}$.
    	\end{enumerate}
    
For the Swiss Roll manifold, the number of samples is 2000 in both experiments. For linear spaces $\mathcal{H}_{3, 30}$ and $\mathcal{H}_{4,30}$ the number of samples is 10000 for the geometrically-probabilistic method, and 1000000 for the Minkowski dimension calculation. 
    
The Swiss Roll manifold is the same as used by Erba \etal as the testing case for the FCI method \cite{Erba}. The linear spaces are analogous to those used in \cite{Erba} but are of smaller dimension. 

	\subsection{Results of the experiments}
	
\begin{figure}[h]
\begin{minipage}{0.45\textwidth}
    \centering\includegraphics[width=0.65\linewidth]{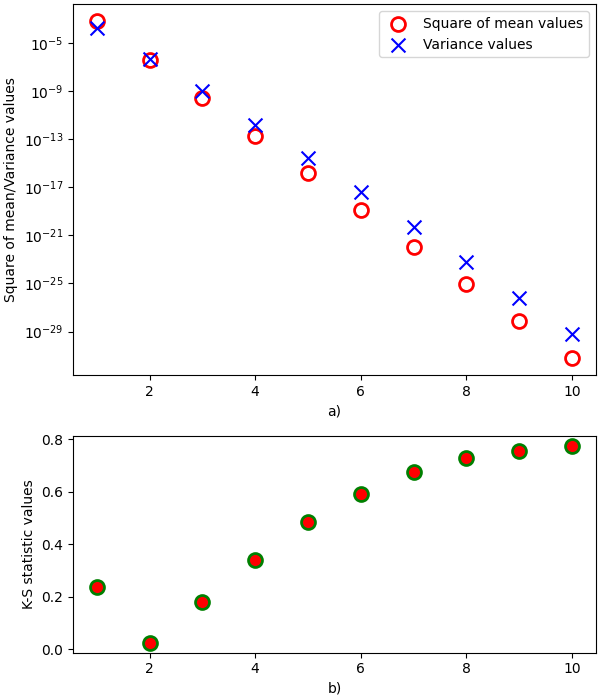}
    \caption{Dataset of 2000 points randomly sampled from the Swiss Roll $\mathcal{SR}_{2,3}$ manifold. a) $A_1^2(n)$ and $A_2(n)$  graphs for different test dimension values $n$; b) K-S statistic graph. The resulting dimension estimate is $n = 2$}
    \label{fig:SR_2000}
\end{minipage}
\hspace{0.05\textwidth}
\begin{minipage}{0.50\textwidth}
    \centering\includegraphics[width=0.9\linewidth]{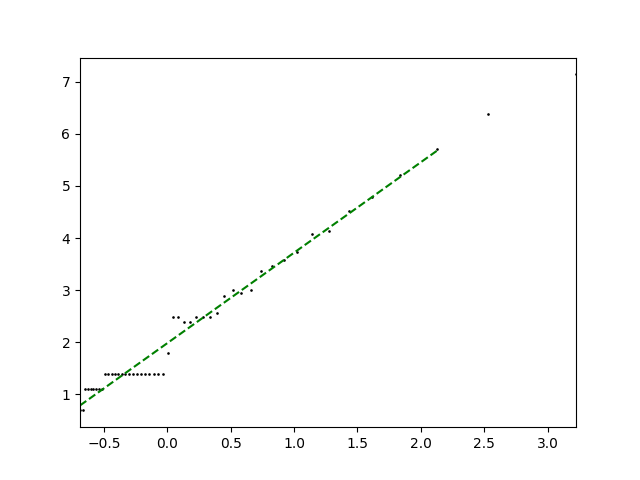}
    \caption{Minkowski dimension estimation for the Swiss Roll dataset. Dimension estimate for $n$ lies between $1$ and $2$} 
    \label{fig:sr_mink}
\end{minipage}
\end{figure}

	Both methods are tested on the Swiss Roll manifold dataset, and the results are coherent. The geometrically-probabilistic approach gives the exact answer of 2, and the Minkowski method slightly underestimates the dimension, giving the approximate answer between 1 and 2 (see Fig.~\ref{fig:SR_2000}, \ref{fig:sr_mink}).
	
\begin{figure}[h!]
\begin{minipage}{0.45\textwidth}
\centering\includegraphics[width=0.65\linewidth]{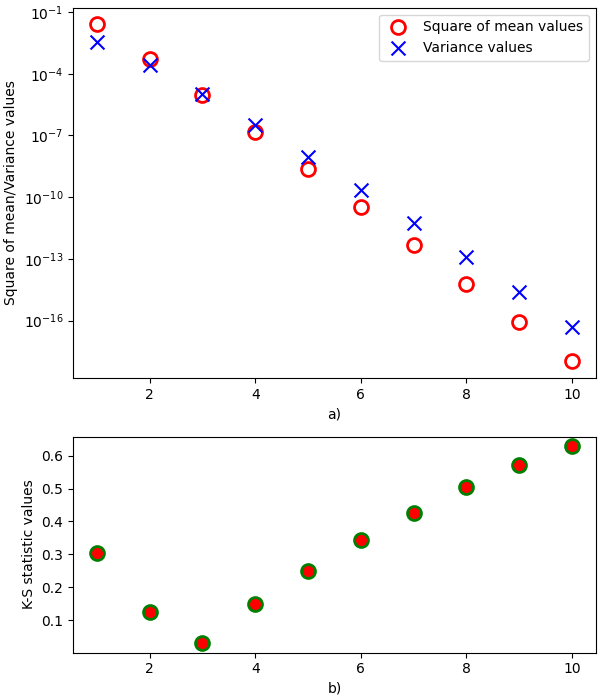}
\caption{$\mathcal{H}_{3,30}$ dataset. a) $A_1^2(n)$ and $A_2(n)$  graphs for different test dimension values $n$; b) K-S statistic graph. The resulting dimension estimate is $n = 3$}
\label{fig:H_3_30}
\end{minipage}
\hspace{0.1\textwidth}
\begin{minipage}{0.45\textwidth}
\centering\includegraphics[width=0.65\linewidth]{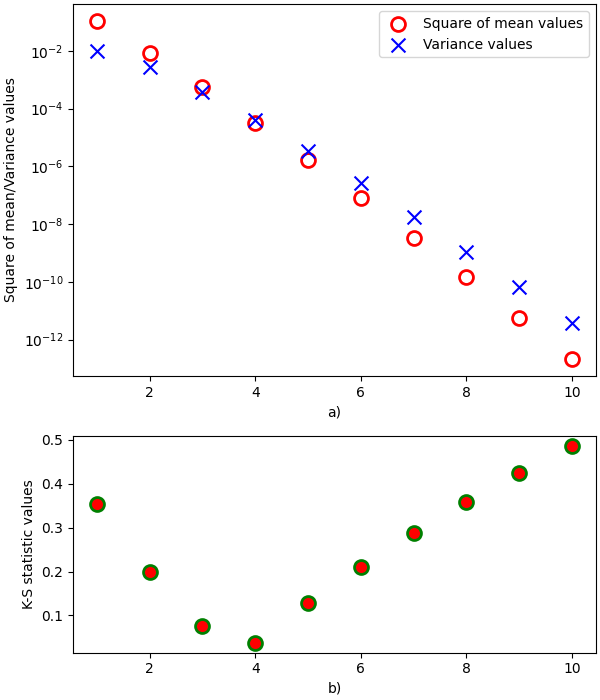}
\caption{$\mathcal{H}_{4,30}$ dataset. a) $A_1^2(n)$ and $A_2(n)$  graphs for different test dimension values $n$; b) K-S statistic graph. The resulting dimension estimate is $n = 4$}
\label{fig:H_4_30}
\end{minipage}
\end{figure}

For the manifolds $\mathcal{H}_{3,30}$ and $\mathcal{H}_{4,30}$ results of the application of the two methods are coherent as well. The dimension estimate given by the geometrically-probabilistic method is exact in both cases: exactly 3 and 4 are obtained (see Fig.~\ref{fig:H_3_30}, \ref{fig:H_4_30}). The Minkowski dimension approach gives the same results: approximately 3 and 4 (see figures in the Supplementary materials to the present paper).
	
\begin{figure}[h!]
\begin{minipage}{0.45\textwidth}
\centering\includegraphics[width=0.8\linewidth]{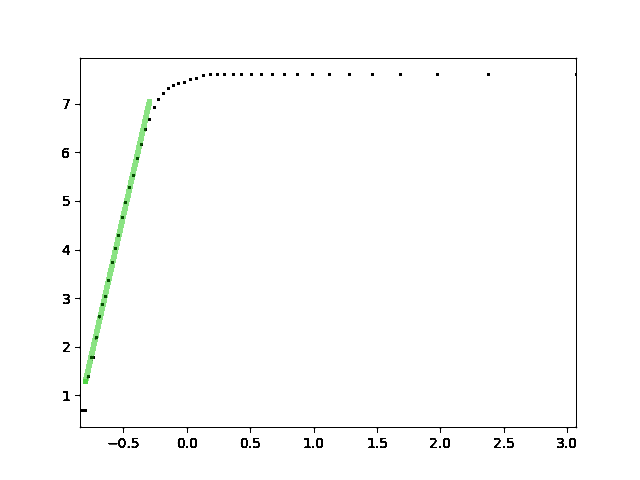}
\caption{Minkowski dimension estimation for the full-faces subset of the set of 11k points. Dimension estimate: $ n \approx 26$}
\label{fig:mink_for11k}
\end{minipage}
\hspace{0.1\textwidth}
\begin{minipage}{0.45\textwidth}
\centering\includegraphics[width=0.65\linewidth]{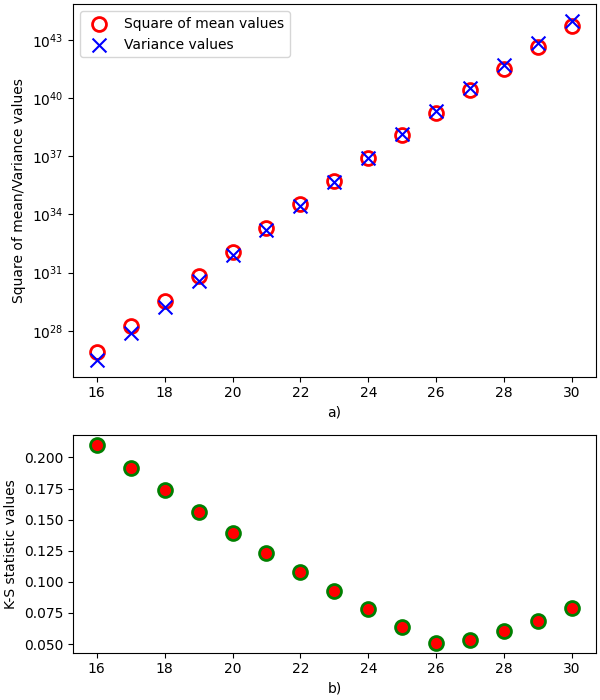}
\caption{Dataset of 3921 points. a) $A_1^2(n)$ and $A_2(n)$  graphs for different test dimension values $n$; b) K-S statistic graph. The resulting dimension estimate is $n = 26$}
\label{fig:dim_for_3921}
\end{minipage}
\end{figure}

	Those results show that for the tested cases the proposed methods give adequate results, correctly estimating the intrinsic dimension. 
	
	The state-of-the-art FCI method due to Erba \etal also estimates the dimension of the Swiss Roll dataset and the linear space datasets correctly, but it is worth noting that our methods have lesser computational complexity than the FCI method.
	
\section{Experiments on Non-Artificial Data}

In the present section we provide several examples of the above-described methods application to certain real-life data clouds. 

\subsection{Experiment setup} 
In our experiments two datasets were used. One was a MS1M-ArcFace dataset (5.8M embeddings of 85K different classes) from {\tt https://github.com/\-deepinsight/insightface/wiki/\-Dataset-Zoo}. These embeddings were obtained by a deep convolutional NN (DCNN) supervised by the ArcFace loss, see \cite{ArcFace}. Another one was a subset of CPLFW dataset (Cross-Pose Labeled Faces in the Wild \cite{CPLFWTech} which is a modified version of a LFW dataset \cite{LFW}). We used a subset that contain 11606 photos of faces including 3921 full faces. To obtain embedding vectors we used DCNN (deep convolutional NN) for face recognition supervised by the ArcFace loss, see \cite{ArcFace}.

For LFW dataset  the following subsets were considered: 

1. $LFW_{3921}$: 3921 embeddings (computed by DCNN from \cite{ArcFace}) for full-faces of different people.  This subset can be considered as  a random sample of independent variables; 

2. $LFW_{11606}$: full dataset which contains 3 images per person, so it has dependencies.

For  MS1M-ArcFace consisting of 5.8M images of people (from \cite{ArcFace}) the following subsets were considered: 

1. $5.8M_{85742}$: a subset  of 85742 images of different people;

2. $5.8M_{\text{first}\, 300}, 5.8M_{\text{middle}\, 300}, 5.8M_{\text{last}\, 300}$: subsets obtained in the following manner. We project the full dataset MS1M-ArcFace to 300 coordinate axis (first 300 axis, 300 axis in the middle, last 300 axis, respectively for the three subsets). Then the Minkowski algorithm was applied to the set of 200000 points extracted from the complete dataset.

\subsection{Results of the experiments}

The set $LFW_{3921}$. In this case --- when dataset consists of independent points --- dimension estimate obtained by geometrically-probabilistic approach ($n = 26$) agrees with dimension estimate obtained by Minkowski method ($n \approx 26$, see Fig.~\ref{fig:mink_for11k}).

Square of mean and variance values and K-S statistic for different test dimensions are shown in Fig.~\ref{fig:dim_for_3921}.

\begin{figure}[h]
\begin{minipage}{0.45\textwidth}
\centering\includegraphics[width=0.65\linewidth]{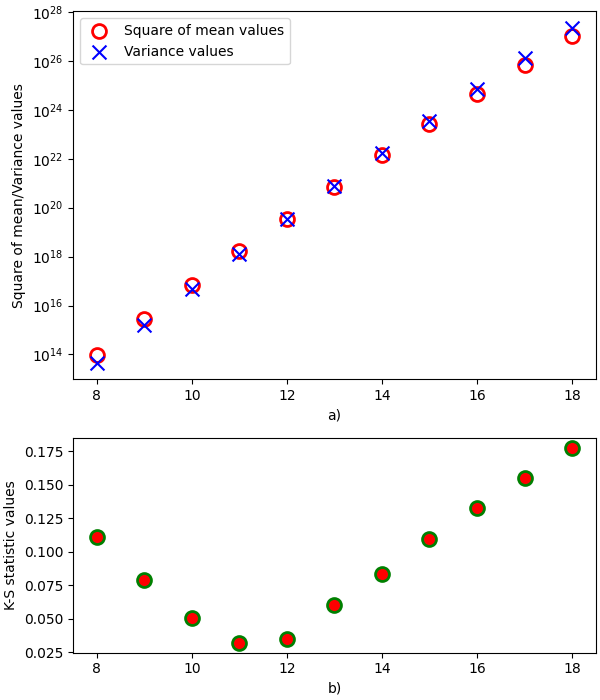}
\caption{Results for LFW. a) Square of mean and variance as functions from test dimension value; b) K-S statistic as function of test dimension value. Due to dependencies dimension is underestimated: $n = 11$ or $12$} 
\label{fig:dim_for_11606}
\end{minipage}
\hspace{0.1\textwidth}
\begin{minipage}{0.45\textwidth}
\centering\includegraphics[width=0.8\linewidth]{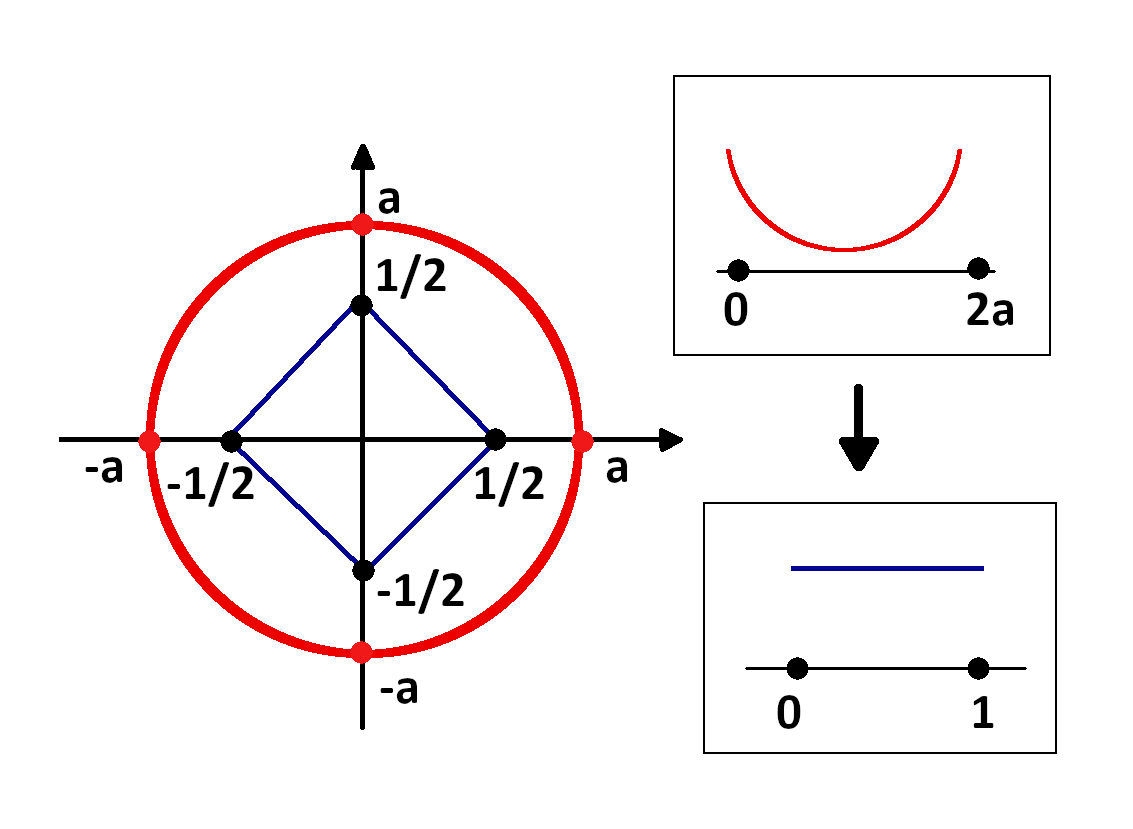}
\caption{Illustration of the flattening procedure on 2D plane. After flattening the circle becomes a rectangle. Note that in high-dimensional case initial red distribution will be quite different, but final blue uniform distribution will be the same}
\label{fig:flattening}
\end{minipage}
\end{figure}

The set $LFW_{11606}$. Here dimension estimate drops to half-value 11-14. It is quite natural: embeddings which refer to the same person should be closer to each other than independent ones. It results in the dimension $n$ being underestimated. Curves for dimension estimation are shown in Fig.~\ref{fig:dim_for_11606}.

\begin{figure}[h]
\begin{minipage}{0.45\textwidth}
\centering\includegraphics[width=0.7\linewidth]{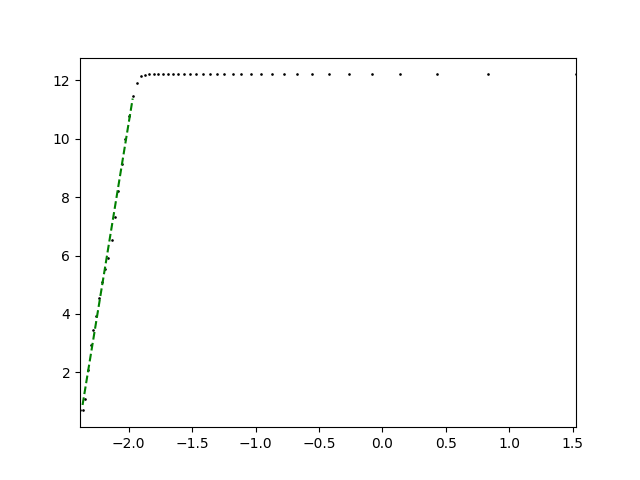} \caption{Minkowski dimension estimation for the projection to the first 300 coordinate axis; 200k points taken with a step of 10. Dimension estimated as 26-27}
\label{fig:middle300_step10}
\end{minipage}
\hspace{0.1\textwidth}
\begin{minipage}{0.45\textwidth}
\centering\includegraphics[width=0.6\linewidth]{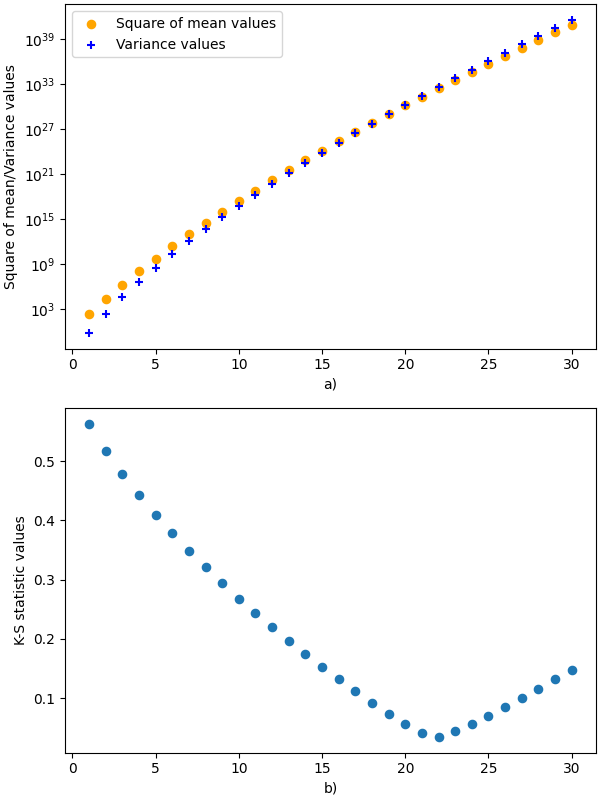}
\caption{Dataset of 85742 points. a) $A_1^2(n)$ and $A_2(n)$  graphs for different test dimension values $n$; b) K-S statistic graph. The resulting dimension estimate is $n = 22$}
\label{fig:dim_for_85742}
\end{minipage}
\end{figure}

The set $5.8M_{85742}$. This dataset was analysed using the geometrically-probabilistic method. The $A_1^2(n), A_2(n)$ and the K-S statistic were considered, and are shown in Fig.~\ref{fig:dim_for_85742}. The intrinsic dimension of the manifold is estimated as $n = 22$.


The sets $5.8M_{\text{first}\, 300}, 5.8M_{\text{middle}\, 300}, 5.8M_{\text{last}\, 300}$. Those subsets were subjected to the Minkowski dimension method. The resulting dimension estimates are very close to each other, which makes a strong argument that those values correctly estimate the dimension of the manifold for the whole data points cloud. Several of the resulting graphics may be seen in Fig.~\ref{fig:middle300_step10} and figures in the supplementary materials to the present paper.
The dimension estimation is $n=25-26$.

Note that the estimates of the dimension of the 5.8M dataset obtained by using different methods (analysing the whole dataset using the statistical method, and splitting the set into parts to apply Minkowski approach to each of them) are coherent.

\begin{figure}[h]
\begin{minipage}{0.45\textwidth}
\centering\includegraphics[width=0.85\linewidth]{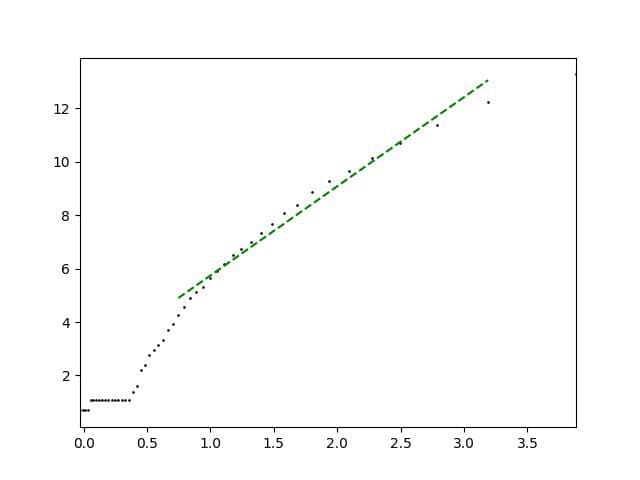}
\caption{Minkowski dimension estimation for the $\mathcal{H}_{3,30}$ dataset. Dimension estimate $n\approx 3$)}
\label{fig:h3_mink}
\end{minipage}
\hspace{0.1\textwidth}
\begin{minipage}{0.45\textwidth}
\centering\includegraphics[width=0.85\linewidth]{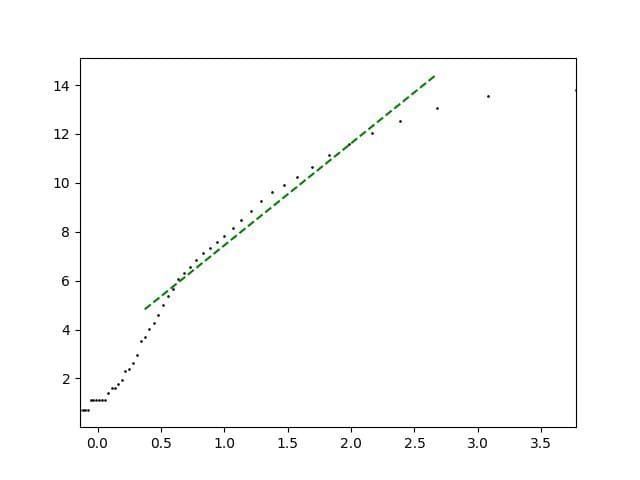}
\caption{Minkowski dimension estimation for the $\mathcal{H}_{4,30}$ dataset. Dimension estimate $n\approx 4$}
\label{fig:h4_mink}
\end{minipage}
\end{figure}

\begin{figure}[h]
\begin{minipage}{0.45\textwidth}
\centering\includegraphics[width=0.85\linewidth]{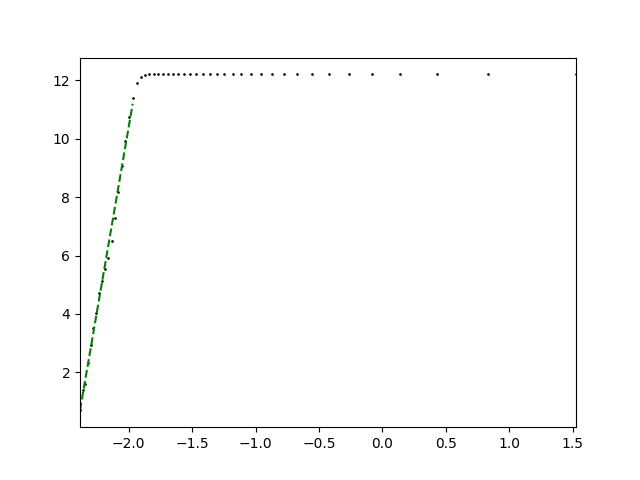} \caption{Minkowski dimension estimation for the projection to the middle 300 coordinate axis; 200k points taken with a step of 10. Dimension estimated as 25-26}
\label{fig:first300_step10}
\end{minipage}
\hspace{0.1\textwidth}
\begin{minipage}{0.45\textwidth}
\centering\includegraphics[width=0.85\linewidth]{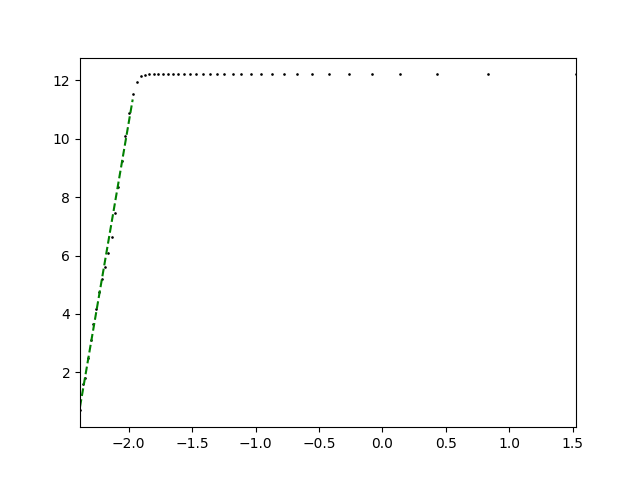} \caption{Minkowski dimension estimation based on 200k points taken with step 10 from dataset $5.8M_{last 300}$. Dimension estimation: 25-26}
\label{fig:last300_step10}
\end{minipage}
\end{figure}

\section{Conclusion} 
We use two methods of manifold dimension estimation which lay upon very different ideas and can be considered as two independent approaches to the same problem. Important fact is that estimates, obtained by them, appear to be very close to each other.  Both methods are lightweight in terms of computations and  allow simple and effective implementation.

\bibliographystyle{plain}
\bibliography{biblio}

\begin{thebibliography}{10}

\bibitem{Burn}
{A. Bernstein, E. Burnaev and P. Erofeev}.
\newblock Manifold reconstruction in dimension reduction problem.
\newblock {\em International conference “Intelligent Information
  Processing” IIP-9}, 9 2012.

\bibitem{EstReach}
Eddie Aamari, Jisu Kim, Frédéric Chazal, Bertrand Michel, Alessandro Rinaldo,
  and Larry Wasserman.
\newblock Estimating the reach of a manifold.
\newblock {\em Electronic Journal of Statistics}, 13, 05 2017.

\bibitem{Anomaly}
{B. Stolz, J. Tanner, H. Harrington and V. Nanda}.
\newblock Geometric anomaly detection in data.
\newblock {\em Proceedings of the National Academy of Sciences}, 117:202001741,
  08 2020.

\bibitem{Feffer3}
{Ch. Fefferman, S. Ivanov, Ya. Kurylev, M. Lassas, and H. Narayanan}.
\newblock {Reconstruction and Interpolation of Manifolds. I: The Geometric
  Whitney Problem}.
\newblock {\em Found. Comput. Math.}, 20, 2015.

\bibitem{Fefferman}
{Charles Fefferman, Sanjoy Mitter, and Hariharan Narayanan}.
\newblock {Testing the manifold hypothesis}.
\newblock {\em J. Amer. Math. Soc.}, 29, 2016.

\bibitem{Feffer2}
{Charles Fefferman, Sergei Ivanov, Matti Lassas, and Hariharan Narayanan}.
\newblock {Reconstruction of a Riemannian Manifold from Noisy Intrinsic
  Distances}.
\newblock {\em SIAM J. on Math. of Data Sc.}, 2, 2020.

\bibitem{AtSoPe}
{Ege Altan, Sara A. Solla, Lee E. Miller, Eric J. Perreault}.
\newblock {Estimating the dimensionality of the manifold underlying
  multi-electrode neural recordings}.

\bibitem{Falconer}
Kenneth {Falconer}.
\newblock {\em {Fractal geometry: mathematical foundations and applications}}.
\newblock Chichester etc.: John Wiley \& Sons, 1990.

\bibitem{LFW}
{G. B. Huang, M. Ramesh, T. Berg, and E. Learned-Miller}.
\newblock Labeled faces in the wild: A database for studying face recognition
  in unconstrained environments.
\newblock Technical Report 07-49, University of Massachusetts, October 2007.

\bibitem{GraCar}
{Granata, D., Carnevale, V.}
\newblock {Accurate Estimation of the Intrinsic Dimension Using Graph
  Distances: Unraveling the Geometric Complexity of Datasets}.
\newblock {\em Sci Rep}, 6, 2016.

\bibitem{Grassm1}
{Ham, Jihun and Lee, Daniel}.
\newblock Grassmann discriminant analysis: a unifying view on subspace-based
  learning.
\newblock {\em Proceedings of the 25th International Conference on Machine
  Learning}, pages 376--383, 07 2008.

\bibitem{ArcFace}
N.~{Xue} J.~{Deng}, J.~{Guo} and S.~{Zafeiriou}.
\newblock Arcface: Additive angular margin loss for deep face recognition.
\newblock In {\em 2019 IEEE/CVF Conference on Computer Vision and Pattern
  Recognition (CVPR)}, pages 4685--4694, 2019.

\bibitem{HeJi}
{J. He, L. Jiang, L. Ding, Z. Ii}.
\newblock {Intrinsic Dimensionality Estimation based on Manifold Assumption}.
\newblock {\em Journal of Visual Communication and Image Representation}, 2014.

\bibitem{Grassm3}
{J. Masci, D. Boscaini, M. Bronstein, and P. Vandergheynst}.
\newblock Geodesic convolutional neural networks on riemannian manifolds.
\newblock {\em {IEEE International Conference on Computer Vision Workshop
  (ICCVW), Santiago, Chile}}, 9:832--840, 2015.

\bibitem{ER}
{J.-P. Eckmann, D. Ruelle}.
\newblock {Fundamental limitations for estimating dimensions and Lyapunov
  exponents in dynamical systems}.
\newblock {\em Physica D: Nonlinear Phenomena}, 56:185--187, 1992.

\bibitem{BeGoRaSha}
{K. Beyer , J. Goldstein , R. Ramakrishnan , U. Shaft}.
\newblock {When is nearest neighbor meaningful}.
\newblock {\em Lect. Notes Comput. Sci.}, 1540:217--235, 1999.

\bibitem{Pearson01}
{Karl Pearson F.R.S.}
\newblock {On lines and planes of closest fit to systems of points in space}.
\newblock {\em {The London, Edinburgh, and Dublin Philos. Magazine and J. of
  Sc.}}, 2, 1901.

\bibitem{MaxLike}
{Levina, E., Bickel, P. J.}
\newblock {Maximum likelihood estimation of intrinsic dimension}.
\newblock {\em Advances in neural information processing systems}, pages
  777--784, 2005.

\bibitem{Kendall}
{M.G. Kendall, P.A.P. Moran}.
\newblock {\em Geometrical probability}.
\newblock Charles Griffin \& Company, London, 1963.

\bibitem{LocPCA}
Kitty Mohammed and Hariharan Narayanan.
\newblock Manifold learning using kernel density estimation and local principal
  components analysis, 09 2017.

\bibitem{Grassberger}
{P. Grassberger, I. Procaccia}.
\newblock Measuring the strangeness of strange attractors.
\newblock {\em {Physica D: Nonlinear Phenomena}}, 9:189--208, 1983.

\bibitem{MorMed}
{P. Mordohai, G. Medioni}.
\newblock {Dimensionality Estimation, Manifold Learning and Function
  Approximation using Tensor Voting}.
\newblock {\em Journal of Machine Learning Research}, 11:411--450, 2010.

\bibitem{Nonlin}
{Roweis, S. and Saul, L.}
\newblock Nonlinear dimensionality reduction by locally linear embedding.
\newblock {\em IEEE International Conference on Computer Vision},
  290:2323--2326, 01 2000.

\bibitem{Riem2}
{Shao, Hang and Kumar, Abhishek and Fletcher, P.}
\newblock The riemannian geometry of deep generative models.
\newblock {\em IEEE/CVF Conference on Computer Vision and Pattern Recognition
  Workshops (CVPRW)}, 11 2017.

\bibitem{CPLFWTech}
{T. Zheng and W. Deng}.
\newblock Cross-pose lfw: A database for studying cross-pose face recognition
  in unconstrained environments.
\newblock Technical Report 18-01, Beijing University of Posts and
  Telecommunications, February 2018.

\bibitem{Erba}
{V. Erba, M. Gherardi, P. Rotondo}.
\newblock {Intrinsic dimension estimation for locally undersampled data}.
\newblock {\em {Sci. Rep.}}, 9, 2019.

\bibitem{dimRed}
Yuanjie Yan, Hongyan Hao, Baile Xu, Jian Zhao, and Furao Shen.
\newblock Image clustering via deep embedded dimensionality reduction and
  probability-based triplet loss.
\newblock {\em IEEE Transactions on Image Processing}, 04 2020.

\end{thebibliography}

\end{document}